\documentclass[conference]{IEEEtran}
\IEEEoverridecommandlockouts
\usepackage{cite}
\usepackage{amsmath,amssymb,amsfonts}
\usepackage{subcaption}
\usepackage[noend, ruled]{algorithm2e}
\usepackage{graphicx}
\usepackage{textcomp}
\usepackage{xcolor}
\usepackage{bm}
\usepackage{makecell}
\usepackage{multirow}
\usepackage{tabularx}
\usepackage{tabulary}
\def\BibTeX{{\rm B\kern-.05em{\sc i\kern-.025em b}\kern-.08em
    T\kern-.1667em\lower.7ex\hbox{E}\kern-.125emX}}
\begin{document}

\title{Multivariate Time Series Classification Using Spiking Neural Networks
}

% \author{\IEEEauthorblockN{Haowen Fang}
% \IEEEauthorblockA{\textit{Department of Electrical Engineering \& Computer Science} \\
% \textit{Syracuse University}\\
% City, Country \\
% email address or ORCID}
% \and
% \IEEEauthorblockN{Amar Shrestha}
% \IEEEauthorblockA{\textit{Department of Electrical Engineering \& Computer Science} \\
% \textit{Syracuse University}\\
% City, Country \\
% email address or ORCID}
% \and
% \IEEEauthorblockN{Qinru Qiu}
% \IEEEauthorblockA{\textit{Department of Electrical Engineering \& Computer Science} \\
% \textit{Syracuse University}\\
% City, Country \\
% email address or ORCID}
% }

\author{\IEEEauthorblockN{Haowen Fang, Amar Shrestha, Qinru Qiu}
\IEEEauthorblockA{Department of Electrical Engineering and Computer Science, Syracuse University, Syracuse, New York\\
Email: { \{hfang02,amshrest,qiqiu\}@syr.edu} }
}

\maketitle

\begin{abstract}
There is an increasing demand to process streams of temporal data in energy-limited scenarios such as embedded devices, driven by the advancement and expansion of Internet of Things (IoT) and Cyber-Physical Systems (CPS). Spiking neural network has drawn attention as it enables low power consumption by encoding and processing information as sparse spike events, which can be exploited for event-driven computation. Recent works also show SNNs' capability to process spatial temporal information. Such advantages can be exploited by power-limited devices to process real-time sensor data. However, most existing SNN training algorithms focus on vision tasks and temporal credit assignment is not addressed. Furthermore, widely adopted rate encoding ignores temporal information, hence it's not suitable for representing time series. In this work, we present an encoding scheme to convert time series into sparse spatial temporal spike patterns. A training algorithm to classify spatial temporal patterns is also proposed. Proposed approach is evaluated on multiple time series datasets in the UCR repository and achieved performance comparable to deep neural networks.
\end{abstract}


\section{Introduction}
The function and behavior of Spiking Neural Networks (SNN) are derived from its inspiration, i.e. the brain, which is capable of performing numerous cognitive tasks with minimal energy requirements. The potential of SNN has drawn various research interests including emerging device, algorithm and applications \cite{yuan2017memristor,shrestha2017spike,9045658,8824944}. In general, the brain's capability comes from the complex dynamics of the networks of spiking neurons and the plastic synapses connecting them. These dynamics can capture complex spatial temporal features of input encoded as sparse temporal spiking activity. Despite the biological inspiration, majority of the existing models of SNNs are unable to replicate such dynamics to encode, learn and decode temporal information.

The limitations of existing SNNs are multifold. Firstly, most SNN models and training algorithms consider only the statistics of spike activities. A numerical value is represented by spike counts in a time window \cite{fang2019general}. Though, this type of SNN has demonstrated state-of-art performance in various tasks\cite{merolla2014million}, it suffers from high spike activities \cite{liu2017mt}. Thus, it cannot fully benefit from event-driven computation. Secondly, directly adapting backpropagation is not feasible because spiking neuron's output is a Dirac delta function. One approach to address the problem is to train an artificial neuron network (ANN) such as multi-layer perceptron (MLP) and map the weights to SNN. However, it suffers form accuracy degradation, additional fine-tuning of weights and thresholds is required to minimize the performance penalty \cite{diehl2015fast}. Recently gradient surrogate is proposed to approximate the gradient of the spiking function, enabling backpropagation  \cite{lee2016training,esser2015backpropagation,shrestha2019approximating,wu2018spatio,shrestha2018slayer,fang2020exploiting}. \cite{esser2015backpropagation} derived a cumulative error function as gradient surrogate. \cite{wu2018spatio} derived a simplified model from Leaky Integrate and Fire neuron (LIF), and proposed four functions as gradient surrogates. Other approaches include replacing hard threshold function by a differentiable soft spike \cite{neftci2019surrogate} \cite{huh2018gradient}. However, it compromises SNN's most distinct feature, binary spike.

While most SNN models and training algorithm use spike counts to resemble numerical value, it is observed that in biological neural networks, temporal structure of spike train also conveys information \cite{butts2007temporal}. Two spike trains of same spike rates can have distinct temporal patterns, hence the represented information is different. Such temporal encoding can efficiently encode information using extremely sparse spikes-events \cite{liu2017mt}. There are some existing works to train neurons to detect temporal spike patterns. Tempotron \cite{gutig2006tempotron} trains individual neuron to perform binary classifications for different spatial temporal input spike patterns. Neuron generate at least one spike for positive pattern, and remain inactive for other patterns. Based on Tempotron, \cite{gutig2016spiking} proposed an algorithm to adjust synaptic weights such that neuron can generate desired number of spikes given a specific input pattern. SPAN \cite{mohemmed2012span} trains an individual neuron to associate a spatial temporal input pattern with a specific output spike pattern. However, these works aim at training individual neurons, cannot be extended to multiple layers, therefore the performance is limited. There are also recent works utilize backpropagation through time (BPTT) to address the temporal dependency problems. \cite{shrestha2018slayer} proposed a training rule to reassign errors in time. \cite{gu2019stca} proposed a novel loss function and derived an iterative model from Tempotron. Based on the iterative model, network can be unrolled hence BPTT is possible. \cite{zenke2018superspike} captures the temporal dependency on membrane potentials and use membrane potential as objective function to learn temporal patterns.

Existing works have achieved comparable performance with DNN in vision tasks such as static image or event-based data classification, however few SNN models address the time series classification tasks. The first challenge is the limitation of rate coding, since it treats spikes statistically, spike coding cannot represent temporal information. Though it is possible to flatten the time series into a 1-D array and then represent it by rate coding, this increases the input size. In addition, for real time applications, flattening input requires buffering, which increases computation latency. Secondly, unlike images, such as MNIST images, where all value's range, precision and scale are identical, multivariate time series may be collected from different sensors, therefore their precision and range may vary. Rate coding has to guarantee the precision for the most high-resolution input. Such "design for worst case" coding scheme lacks flexibility and hence is not efficient. New coding methods to represent time series are required to exploit the potential of SNNs.

In this work, our contributions are summarized as follows:

\begin{itemize}
    \item We present a coding method that can convert time series into sparse spatial temporal spike patterns. 
    \item We derive an iterative SNN model from Spike Response Model, such that the Backpropagation Through Time is possible. An event-based updating algorithm is also proposed to reduce computation overhead for inference.
    \item We formulate a backpropagation rule for the iterative SNN model and propose a training algorithm to train the model on spatial temporal patterns.
    \item We evaluate the proposed method on multiple datasets, and achieve comparable accuracy with DNN. To the best of our knowledge, this is the first work applying SNN for multivariate time series classification.
\end{itemize}

\section{SNN Model}

Without loss of generality, we adopt the a widely used Leaky Integrate and Fire (LIF) neuron defined by Spike Response Model \cite{gerstner2002spiking,gutig2006tempotron}. Each input spike induces a charge in the neuron's membrane potential, which is called a postsynaptic potential (PSP):
\begin{equation}\label{eq:psp}
PSP(t) = \sum_{t_i}^{t_i < t}{K(t-t_i)}w
\end{equation}
where $t_i$ denotes the arrival time of $i_{th}$ input spike. $K(t)$ is synapse kernel. Neuron accumulate all input PSPs, such that the membrane potential $v(t)$ is defined as:
\begin{equation}\label{eq:srm}
v(t) = \sum_{i}^{N}{w_i}PSP_i(t) - V_{th}\sum_{t_s^j<t} e^{-\frac{t-t_s^j}{\tau}} 
\end{equation}

\begin{figure}
  \centering
  \includegraphics[width=\linewidth]{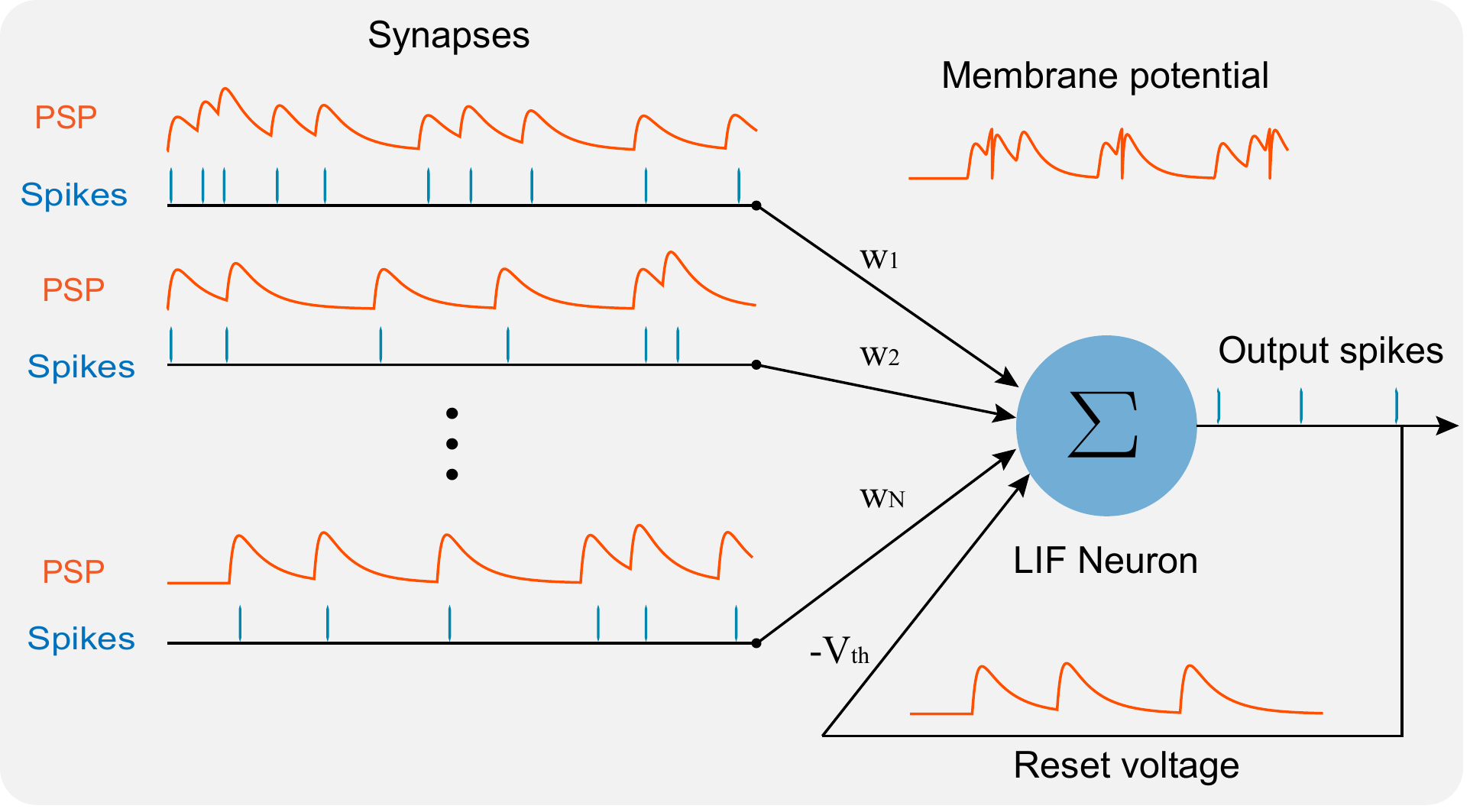}
  \caption{Spiking neuron model}
  \label{fig:model}
\end{figure}
\begin{figure}
  \centering
  \includegraphics[width=\linewidth]{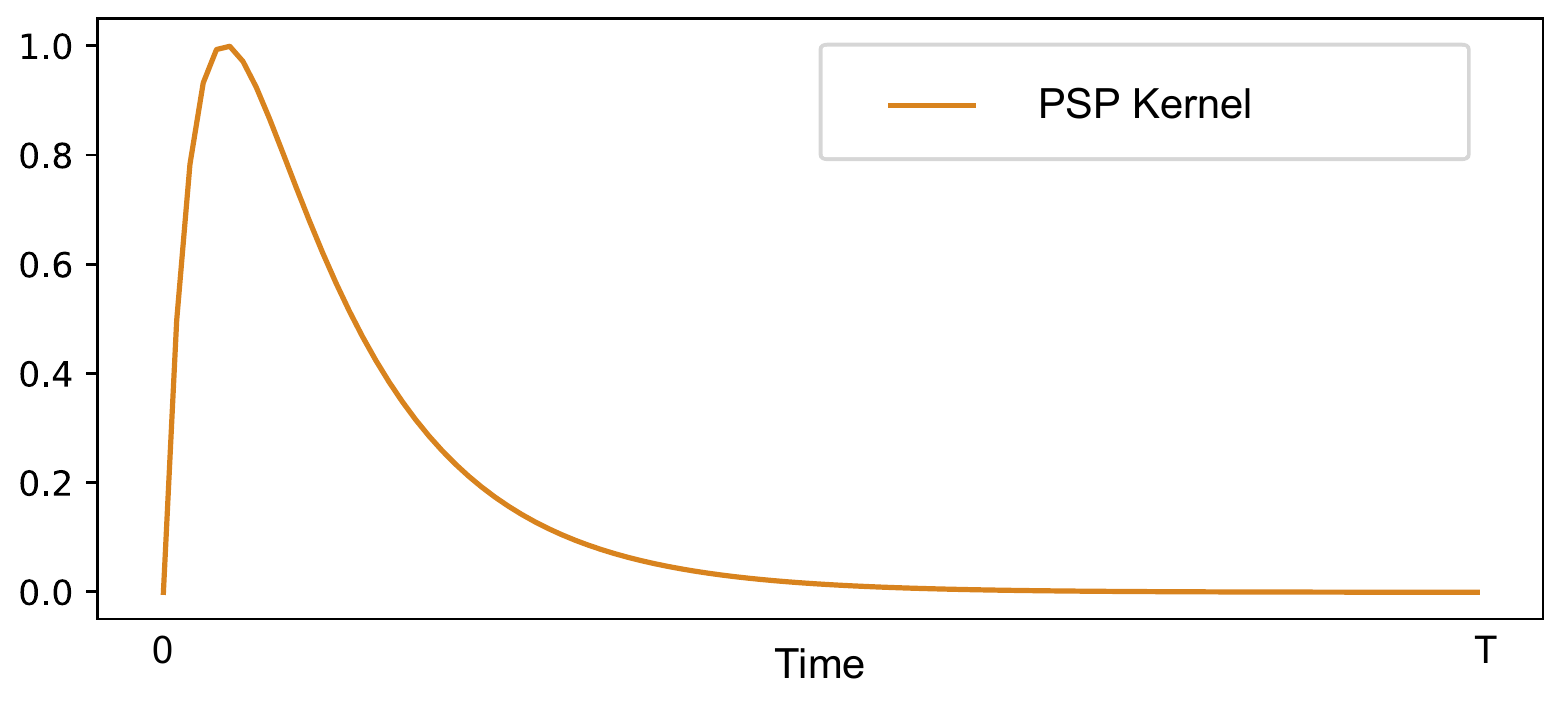}
  \caption{PSP kernel}
  \label{fig:kernel}
\end{figure}


where $N$ is the number of input synapse. $t_j^i$ is the arrival time of $j_{th}$ spike at $i_{th}$ input synapse. $\tau$ is a time constant of neuron. $w_i$ is the weight associated with each input synapse. $t_s^j<t$ is the time when the neuron generates an output spike and the rightmost term can be interpreted as a negative voltage applied to the neuron itself such that the membrane potential is decreased by a factor of the threshold voltage $V_{th}$. This serves as the reset mechanism at the time of spike. Thus, the neuron’s potential is the summation of all weighted input PSP plus the negative voltage given by rightmost term \cite{gerstner2002spiking}. The neuron model with $N$ inputs is illustrated in figure \ref{fig:model}. PSP kernel $K(t)$ is defined as:
\begin{equation}\label{eq:Kernel}
K(t) = V_0 (e^{-\frac{t}{\tau_m}} - e^{\frac{-t}{\tau_s}})
\end{equation}
where $\tau_m$ and $\tau_s$ are two time constants. $V_0 = \eta^{\frac{\eta}{\eta - 1}}$ is a normalization factor which scales the maximum value of $K(t)$ to 1, and $\eta = \frac{\tau_m}{\tau_s}$ \cite{gutig2016spiking}. The shape of the PSP kernel is shown in figure \ref{fig:kernel}.

 At time $t$, PSP and membrane potential are determined by all previous inputs. However, it is not feasible to directly implement the SNN model defined by Equation \ref{eq:srm}. At any given time $t$, $v(t)$ has to be computed by recursively convolving input spike trains with $K(t)$, thus, incurring significant computation overhead. To address this issue, an incremental way to update the PSP can be derived from the SRM model in discrete time domain.

More formally, input spike train $S[t]$ can be defined as a sequence of time shifted Dirac Delta functions:
\begin{equation}\label{eq:input}
S[t] = \sum_n^t x[n]\delta[t-n] 
\end{equation}
where $x_i[t] = 1$ denotes a spike received at time $t$, otherwise $x_i[t] = 0$. Similarly, output spike train $O[t]$ can defined as:
\begin{equation}\label{eq:output}
O[t] = \sum_n^t y[n]\delta[t-n] 
\end{equation}
where $y[t]$ satisfies $(v[t] < V_{th} \rightarrow y[t] = 0) \cap (v[t] > V_{th} \rightarrow y[t] > 0)$. To derive the incremental model, We define $M[t] = \sum_n e^{-\frac{n}{\tau_m}}S[t-n]$, $H[t] = \sum_n e^{-\frac{n}{\tau_s}}S[t-n]$, such that the $PSP$ can be expressed as the combination of $M[t]$ and $H[t]$ \cite{yu2018spike}:
\begin{equation}\label{eq:psp_sum}
PSP[t] = V_0(M[t] - H[t])
\end{equation}
$M[t]$ and $H[t]$ can be computed incrementally:
\begin{equation}\label{eq:N}
M(H)[t] = e^{\frac{-1}{\tau_m(s)}} M(H)[t-1] + S[t]
\end{equation}
Similarly, we can compute reset voltage $R[t]$:
\begin{equation}\label{eq:R}
R[t] = e^{\frac{-1}{\tau_s}} R[t-1] +  O[t-1]
\end{equation}
Such that the SNN defined in Equation \ref{eq:srm} can be equivalently expressed as:
\begin{subequations}
\label{eq:neuron_model}
\begin{gather}
      V^l_i[t] = I^l_i[t] - V_{th} R^l_i[t]       \label{eq:neuron_model_1} \\
      I^l_i[t] = V_0 \sum_j^{M_{l-1}} w^l_{i,j} (M^l_i[t] - H^l_i[t])       \label{eq:neuron_model_2} \\
      M^l_j[t] =  \alpha N^l_j[t-1] +  O^{l-1}_j[t]  \label{eq:neuron_model_3} \\
      H^l_j[t] =  \beta H^l_j[t-1] +  O^{l-1}_j[t] \label{eq:neuron_model_4} \\
      R^l_i[t] = \gamma R^l_i[t-1] +  O^l_i[t-1] \label{eq:neuron_model_5} \\
      O^L_i[t] = U(V^l_i[t]-V_{th}) \label{eq:neuron_model_6}
\end{gather}
\end{subequations}

where indexes $l$, $i$, $j$ denote layer index, neuron index and input index respectively. $N_l$ denotes the number of neurons in $l_{th}$ layer. $I^l_i[t]$ is input current, $R[t]$ is reset voltage, and $O^l_i[t]$ is neuron output. $\alpha = e^{\frac{-1}{\tau_m}}$, $\beta = e^{\frac{-1}{\tau_s}}$, $\gamma = e^{\frac{-1}{\tau}}$ are three decay factors. More specifically, $l=0$ denotes the encoding layer, which will be discussed in section \ref{sec:coding}, $L$ is the number of layers in the network and $l = L$ denotes output layer. $U(x)$ is a Heaviside step function:
\begin{equation}\label{eq:step_function}
U(x) = 0, \text{if } x < 0 \text{, otherwise 1}
\end{equation}
In the above model, the temporal dependency can be clearly seen in Equation \ref{eq:neuron_model_1} - \ref{eq:neuron_model_6}. At each time $t$, the PSP, membrane potential and output can be computed based on time $t - 1$, hence by unfolding the network, Backpropagation Through Time (BPTT) can be used to train the network. Note that the gradient of $U(x)$ is a Dirac Delta function, therefore backpropagation cannot be directly applied. Its approximation will be discussed in \ref{sec:training}.

\subsection{Event-driven Inference}
The above model provides an explicit approach to update SNN's states based on step-wise computation, and is suitable for training. In inference, the model can be simulated in an event-driven manner, i.e. computation is only necessary when there is a spike event, hence significantly reducing the computation overhead.

Suppose at time $t$, the value of $M[t]$ or $H[t]$ is known, without input spike, after $\Delta t$ unit time later, i.e. at time $t' = t + \Delta t$, the $M[t]$ and $H[t]$ can be computed as:
\begin{equation}\label{eq:exp_iter}
\begin{split}
M(H)[t'] &= \sum_{t_i}^{t_i < t}  e^{-\frac{t+\Delta t-t_i}{\tau_{m(s)}}} \\
                  &=\sum_{t_i < t'} e^{-\frac{t-t_i}{\tau_{m(s)}}} \cdot e^{-\frac{\Delta t}{\tau_{m(s)}}}\\
                  &= M(H)[t] e^{\frac{-\Delta t}{\tau_{m(s)}}}
\end{split}
\end{equation}

When there is an input spike at time $t' = t + \Delta t$. There is an instantaneous unit charge on $M[t']$ and $H[t']$:
\begin{equation}\label{eq:exp_update}
\begin{split}
M(H)(t') &= M(H) e^{\frac{-\Delta t}{\tau_{m(s)}}} + 1
\end{split}
\end{equation}

Similar update rule applies for $R[t]$:
\begin{equation}\label{eq:reset}
\begin{split}
R(t+\Delta t) = R[t]e^{\frac{-\Delta t}{\tau_m}}
\end{split}
\end{equation}

\begin{algorithm} [t]
Input spike buffer: $Q_{spike} \leftarrow \emptyset$\\
Elapsed time since last input spike: $D_{in}[:] \leftarrow 0$\\
Elapsed time since last output spike: $D_{out}[:] \leftarrow 0$\\
Time $t \leftarrow 0$ \\
$M[:] \leftarrow 0$ \\
$H[:] \leftarrow 0$ \\
$R[:] \leftarrow 0$ \\

\For{$t < T$}
{
    \eIf{$Q_{spike} \neq \emptyset$}
    {
    \ForEach{synapse $j$}
    {
        $V \leftarrow 0$ \\
        \eIf{$j$ in $Q_{syn}$}
        {
            $M[j] \leftarrow M[j] \cdot \alpha^{D_{in}[j]} + 1$ \\
            $H[j] \leftarrow H[j] \cdot \beta^{D_{in}[j]} + 1$ \\
            $D_{in}[j] \leftarrow 0$ \\
            $V \leftarrow V + V_0 \cdot w_j \cdot (M[j] - H[j])$\\
        }
        {
            $D_{in}[i] \leftarrow D_{in}[i] + 1$ \\
        }
    }
        \eIf{$V > V_{th}$}
        {
            $D_{out} \leftarrow0$ \\
            $R \leftarrow R \cdot \gamma^{D_{out}[j]} + V_{th}$ \\
            $V \leftarrow V - V_{th}$
        }
        {
            $D_{out} \leftarrow D_{out} +1$
        }
     }
    {
        $D_{in} \leftarrow D_{in} + 1$ \\
        $D_{out} \leftarrow D_{out} + 1$ \\
    }
    $t \leftarrow t+1$
}

 \caption{Event-driven inference}
 \label{alg:event_update}
\end{algorithm}

The event-driven computation algorithm is shown in Algorithm \ref{alg:event_update}. By tracking the elapsed time $\Delta t$, computation is only necessary when there is an input or output spike. In addition, the kernel decays over time, and becomes effectively 0 after a period. Therefore the decay factor for different $\Delta t$ can be pre-computed and stored in a look-up table, so that the expensive exponential function is avoided.

\section{Spatial Temporal Population Encoding} \label{sec:coding}
Rate coding represents a numerical value by the activity of an individual neuron that fires at a particular rate. For example, in vision tasks, to encode one pixel, a spike train's spike count $C$ in a given time window $T$ is proportional to the pixel value. There are several drawbacks of rate coding. 1) the precision is limited because the value represented by rate coding is quantized by bin size $1/T$. Though higher precision can be obtained by increasing $T$, the computational latency increases as well; 2) it is unable to represent temporal information as it treats spike activity statistically. Time series have to be flattened and then converted to spike trains. In real time scenarios, it requires data stream to be buffered, which causes additional latency; 3) Individual neuron is too noisy due to stochastic nature, thus it introduces additional noise; 4) it causes high spiking activity, as larger value has to be represented by more spikes, which deprives the SNN energy efficiency. 5) It is incapable of representing negative values, which are common in sensor inputs.

To address the above issues, we employ a coding method suitable for encoding time series by combining population coding and temporal coding \cite{fang2019event}. In population coding the information is represented by the activity of a group of neurons. Inside a population, each neuron has its favorable input, i.e. each neuron responds to a particular input and remains relatively inactive for other inputs. In temporal coding, the spike train patterns also convey information.

We utilize a population of Current-based Integrate and Fire(CUBA) neurons as encoder. A CUBA neuron is defined as a hybrid system \cite{brette2007simulation}:
\begin{equation}\label{eq:cuba}
\begin{split}
\frac{dV}{dt} &= -\frac{1}{\tau}V + g \cdot I_{ext}(t) \\
            V &\leftarrow 0  \qquad \text{if $V > V_{th}$}
\end{split}
\end{equation}

where $I_{ext}(t)$ is the external time-varying input current, $\tau$ is the membrane time constant, which determines the decay speed of membrane potential, $V(t)$ is the neuron membrane potential, $g$ is the gain. Neuron accumulates the input current and updates the membrane potential continuously. When the $V(t)$ exceeds $V_{th}$, a reset is triggered, membrane potential is forced to 0.

In practice, CUBA neuron model is simulated in discrete time, $V(t)$ is evaluated on a time grid and the interval of the grid is $dt$ \cite{rotter1999exact}. $I_{ext}(t)$ is also sampled at each time grid. Such that the discrete version of the model represented by Equation \ref{eq:cuba} is:
\begin{equation}\label{eq:discrete_cuba}
\begin{split}
V[t+1] &= e^{-\frac{dt}{\tau}} V[t] + g \cdot I_{ext}[t] \\
            V[t+1] &= 0  \qquad \text{if $V[t] > V_{th}$}
\end{split}
\end{equation}

\begin{figure}
  \centering
  \includegraphics[width=0.9\linewidth]{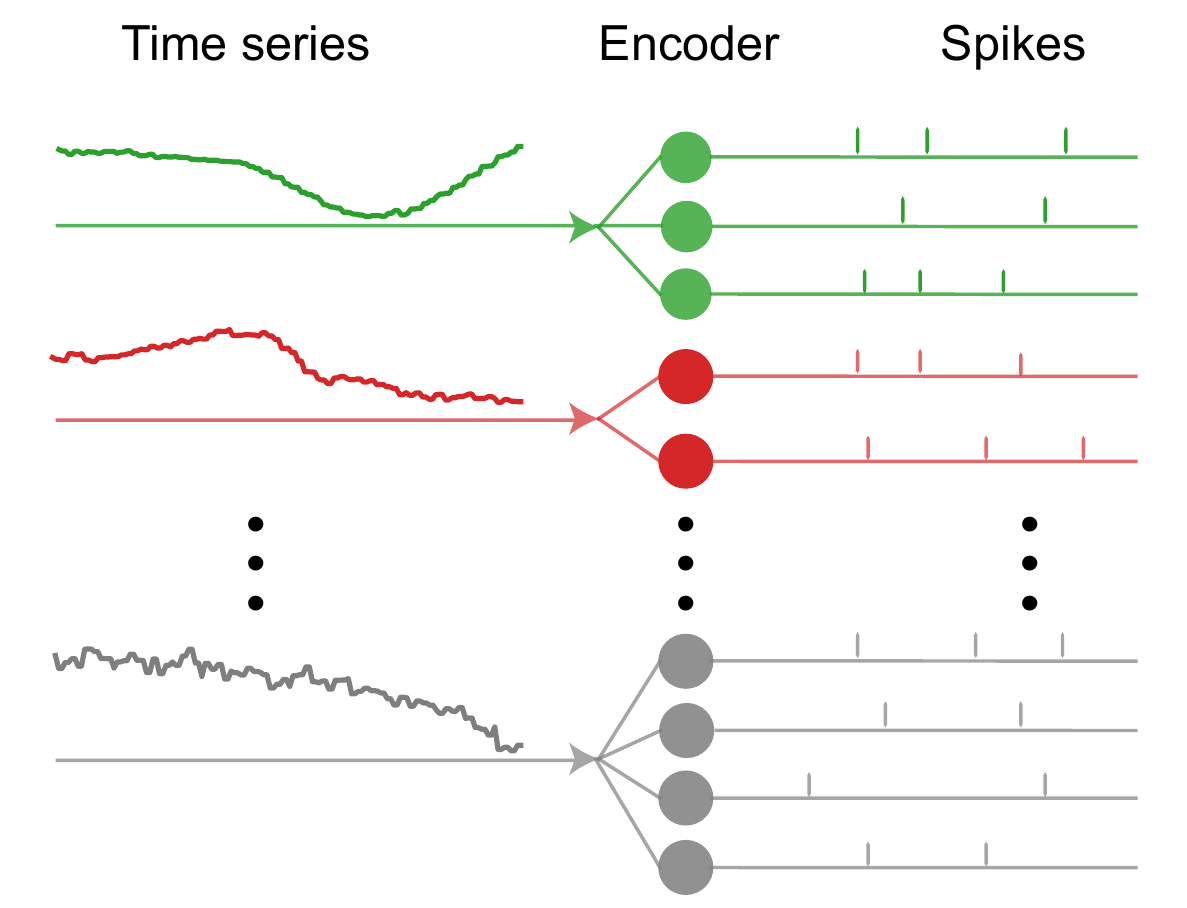}
  \caption{Coding method}
  \label{fig:coding}
\end{figure}

With this coding scheme, a univariate time series, for example sensor data is treated as input current and connected to a population of $E$ encoder neurons. Each neuron may have different time constant $\tau$ and gain $g$, so that each encoder responds to the input differently. In addition, by setting $g$ to negative, neuron can also respond to negative input, which overcomes the drawback of rate coding. We utilize Neural Engineering Framework (NEF)\cite{eliasmith2004neural} to pre-train the encoder. Using this approach, time varying signal is converted into time varying spike patterns. For multivariate time series of $C$ channels, each channel can be encoded using the above approach. Unlike vision tasks in which all input dimensions have identical resolution and range, multivariate time series maybe collected from different sensors, therefore the resolution, precision and range may vary. By coding method, the population size and tuning curve can be adjusted to provide just enough precision for all the channels \cite{yarrow2015influence,fang2019event}.

\section{Training Algorithm}\label{sec:training}

For the classification task, the neuron in the output layer that fires most frequently represents the result, we use cross-entropy loss defined as:
\begin{equation} \label{eq:loss}
E = -\sum_i^{N_L} y_i log(p_i)
\end{equation}
where $p_i$ is the probability of each class calculated by softmax, $p_i$ given by:
\begin{equation} \label{eq:p}
p_i = \frac{\exp{(\sum_t^T O^L_i[t]})}{\sum_{j=1}^{N_L} \exp{(\sum_t^T O^L_j[t])}}
\end{equation}

where $y_i$ is the label, $L$ is number of layers, $O^L_i[t]$ denotes output of last layer, and $N_L$ is the neuron number of last layer.

Equation \ref{eq:neuron_model_1} - \ref{eq:neuron_model_6} provide an explicit way to update SNN states and outputs. By unfolding the network, BPTT can be used for training. First, we define $\delta^{l}_i[t] = \frac{{\partial L}}{{\partial O^l_i[t]}}$, $\epsilon^{l}_i[t] = \frac{{\partial U(V^{l}_i[t]-V_{th})}}{{\partial V^{l}_i[t]}}$ and $\kappa^l_i[t] = \frac{\partial O^l_i[t+1])}{\partial O^l_i[t]}$. 

$\kappa^l_i[t]$ is given by:
\begin{equation} \label{eq:kappa}
\kappa^l_i[t] = -V_{th} \gamma \epsilon^l_i[t]
\end{equation}

At last layer $L$, $\delta^L_i[t]$ can be computed as:
\begin{align} \label{eq:last_delta}
\delta^L_i[t] &= \frac{\partial E}{\partial O^L_i[k]} = \frac{\partial E}{\partial (\sum_{k=1}^T O^L_i[t])} \frac{\partial (\sum_{k=1}^T O^L_i[t])}{\partial O^L_i[t]} \nonumber \\
& = (p_i - y_i)(\sum_{k=t}^{T} \frac{\partial O^L_i[k]}{\partial O^L_i[t]})
% & = (p_i - y_i)(1 + \sum_{k=1}^T\prod_{n=1}^k \kappa^l_i[t+n]\delta^l_i[t+n] )
\end{align}
$\frac{\partial O^L_i[k]}{\partial O^L_i[t]}$ is computed as:
\begin{align}  \label{eq:o_grad}
\frac{\partial O^L_i[k]}{\partial O^L_i[t]} = \prod_{n=0}^{k-t-1} \frac{\partial O^L_i[t+n+1]}{\partial O^L_i[t+n]} = \prod_{n=0}^{k-t-1}(-V_{th}\gamma \epsilon^L_i[t+n])
\end{align}

For hidden layer $l < L$, $\delta^l_i[t]$ can be computed recursively from output layer $L$ and time $T$ to input layer and time 0:
\begin{align} \label{eq:delta}
\delta^{l,i}[t] &= \sum_{j}^{N_{l+1}} \frac{\partial E}{\partial O^{l+1}_j[t+1]}\frac{\partial O^{l+1}_j[t]}{\partial O^l_i[t]} \nonumber + \frac{\partial E}{\partial O^l_i[t]}\frac{\partial O^l_i[t+1]}{\partial O^l_i[t]} \nonumber \\&
= -V_{th}\delta^l_i[t+1]\epsilon^l_i[t+1] \nonumber \\ &+ \sum_{j}^{N_{l+1}} w_j^i \delta^{l+1}_i[t+1]\epsilon^{l+1}_i[t+1](\alpha - \beta)
\end{align}

Heaviside step function $U(x)$ is non-differentiable. We employ gradient surrogate \cite{neftci2019surrogate} to address this issue. In forward path, the spike generation mechanism remains unchanged, while in the backward path, the derivative of $U(x)$ is replaced by the derivative of a smooth function. We use a sigmoid function proposed by \cite{wu2018spatio} as the gradient surrogate in the backward path, such that the gradient of $U(x)$ is approximated as:
\begin{equation}
\frac{\partial U(v)}{\partial V} \approx \frac{e^{V_{th} - v}}{(1 + e^{V_{th} - v})^2}
\end{equation}

Based on above equations, the gradient of weight can be computed as:
\begin{align}
\frac{\partial E}{\partial\mathbf{w}^l} &= \sum^T_{t=1} \frac{\partial E}{\partial \bm{O}^l[t] } \frac{\partial \bm{O}^l}{\partial \bm{V}^l[t] } \frac{\partial \bm{V}^l[t]}{\partial \bm{I}^l[t] } \frac{\partial \bm{I}^l[t]}{\partial\mathbf{w}^l } \nonumber \\&
= \sum^T_{t=1} V_0 \cdot \bm{\delta}^l[t] \bm{\epsilon}^l[t](\bm{M}^l[t] - \bm{N}^l[t])
\end{align}

\begin{algorithm} [t]
Input: Time-varying input $\bm{I}_{ext}[t]$ \\
Output: Optimized weights $\bm{W}^l$ \\
\tcp{Forward}
\For{$t=1$ to $T$}
{   
    \tcp{Encoding}
    $\bm{V}^0 \leftarrow \bm{e}^{\frac{-1}{\tau}} \cdot \bm{V}^0 + \bm{I}_{ext}[t]$ \\
    \eIf{$\bm{V}^0 > V_{th}$}
    {
        $\bm{V}^0 \leftarrow 0$ \tcp{Reset} 
        $\bm{O}^0 \leftarrow 1$ \tcp{Generate spike}
    }
    {
        $\bm{O}^0 \leftarrow 0$ 
    }
    \For{$l=1$ to $L$}
    {
    \tcp{Update states Eq.\ref{eq:neuron_model_1} -\ref{eq:neuron_model_5}}
    $(\bm{M}^l, \bm{H}^l, \bm{R}^l, \bm{V}^l, \bm{I}^l) \leftarrow \text{Update}(\bm{M}^l, \bm{H}^l, \bm{R}^l, \bm{O}^{l-1}, \bm{O}^{l})$
    $\bm{O}^{l} \leftarrow \text{SpikeFunction}(\bm{V}^{l})$ \tcp{Eq.\ref{eq:neuron_model_6}}
    }
    
    \tcp{Calculate loss}
    $E = \text{Loss}(\bm{O}^L[1], ..., \bm{O}^L[T])$ \tcp{Eq.\ref{eq:loss}-\ref{eq:p}}
    
     \tcp{Backward}
}   
 \For{$t = T$ to $1$}
 {  
    $(\bm{\delta}^L[t-1], \bm{\kappa}^L[t-1]) \leftarrow \text{BackProp}(\bm{E}, \bm{\delta}^L[t], \bm{\kappa}^L[t])$ \tcp{Eq.\ref{eq:last_delta}-\ref{eq:o_grad}}
    
    \For{$l = L$ to $1$}
    {
    $(\bm{\delta}^{l-1}[t-1], \bm{\kappa}^{l-1}[t-1]) \leftarrow \text{BackProp}(\bm{\delta}^l[t], \bm{\kappa}^l[t])$ \tcp{Eq.\ref{eq:kappa},\ref{eq:delta}} 
    }
 }
 \caption{Training process of one iteration}
 \label{alg:training}
\end{algorithm}

\section{Experiments}
The proposed network model and algorithm are implemented in PyTorch. We demonstrate the effectiveness of our work in two experiments. In the first experiment, we compared the coding efficiency of rate coding and temporal population coding in terms of spike rate and input size. In second experiments, the proposed network and algorithm is evaluated on various multivariate time series classification tasks.

\subsection{Coding Efficiency}

\begin{figure}
  \centering
  \includegraphics[width=\linewidth]{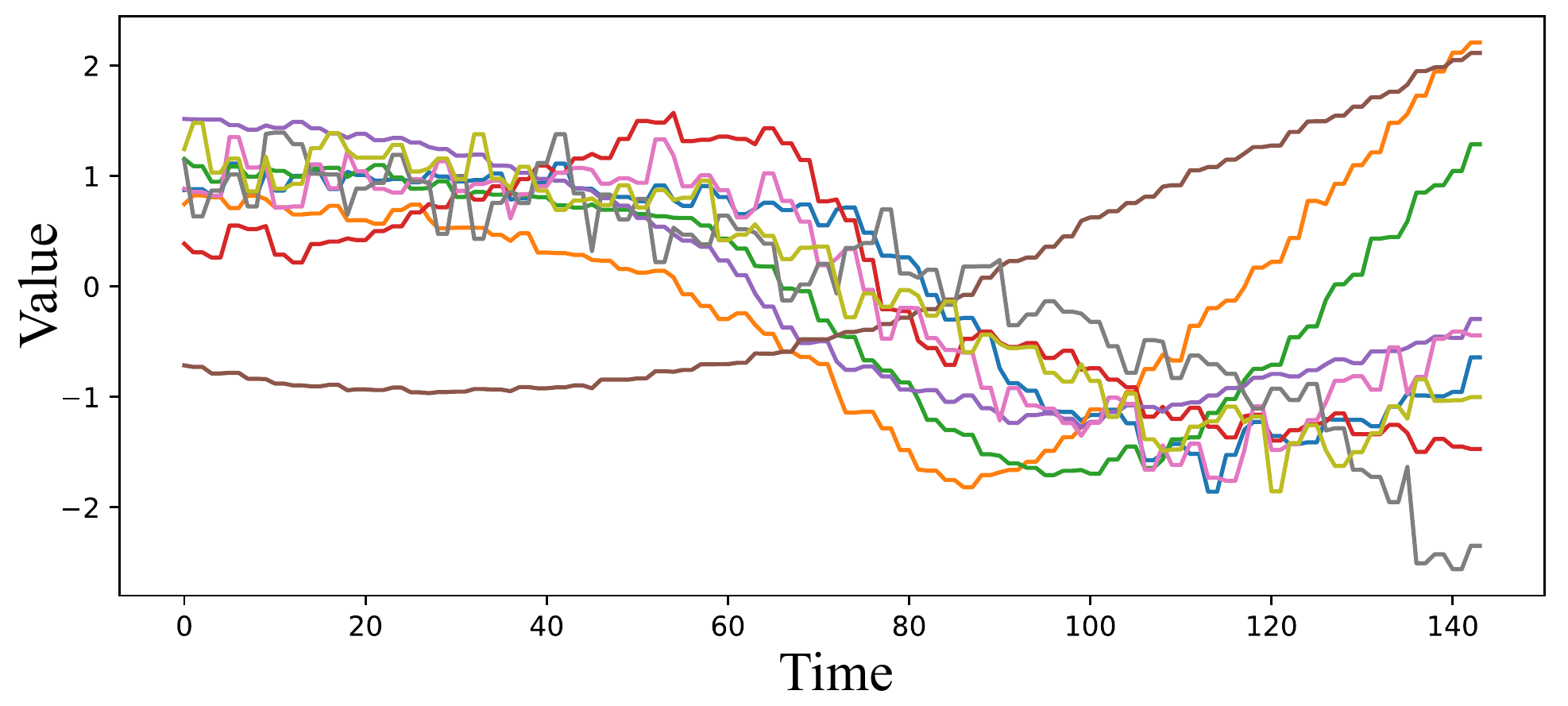}
  \caption{Articulary Word Recognition dataset sample}
  \label{fig:sample}
\end{figure}

\begin{figure}[ht]
  \centering
    \begin{subfigure}{\linewidth}
    \includegraphics[width=0.9\linewidth]{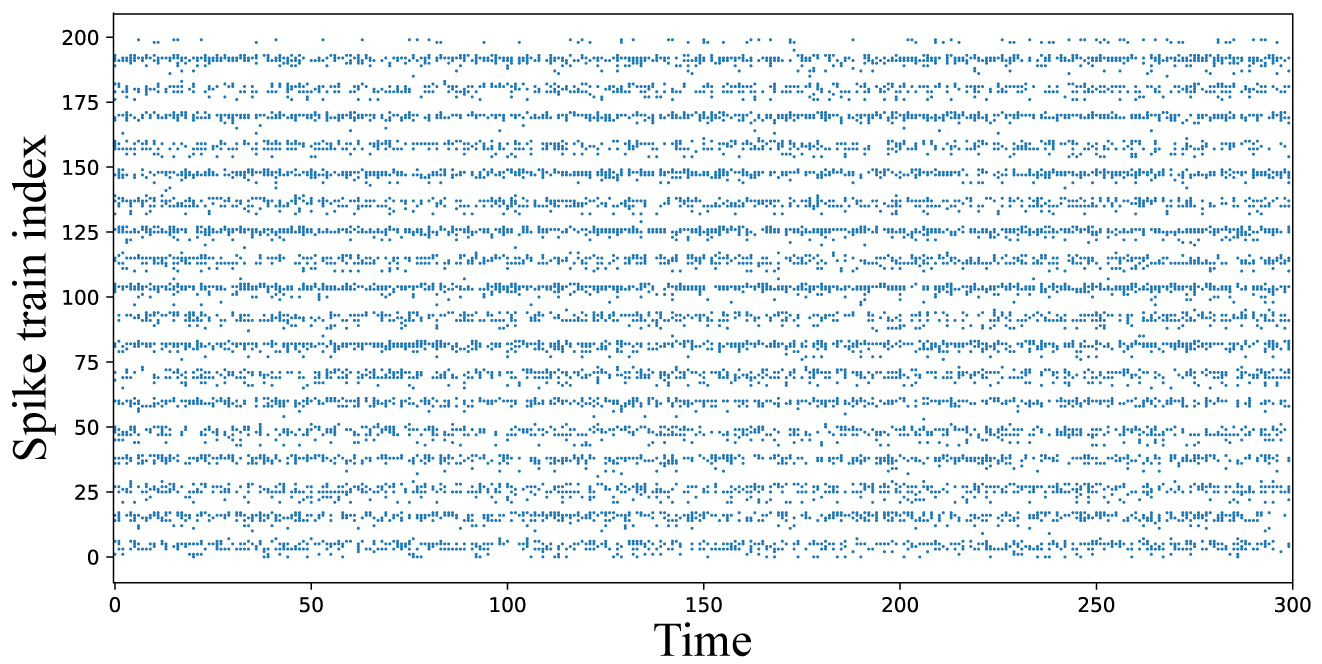}
    \caption{Rate coding}\label{fig:rate}
  \end{subfigure}
  
  \begin{subfigure}{\linewidth}
    \includegraphics[width=0.9\linewidth]{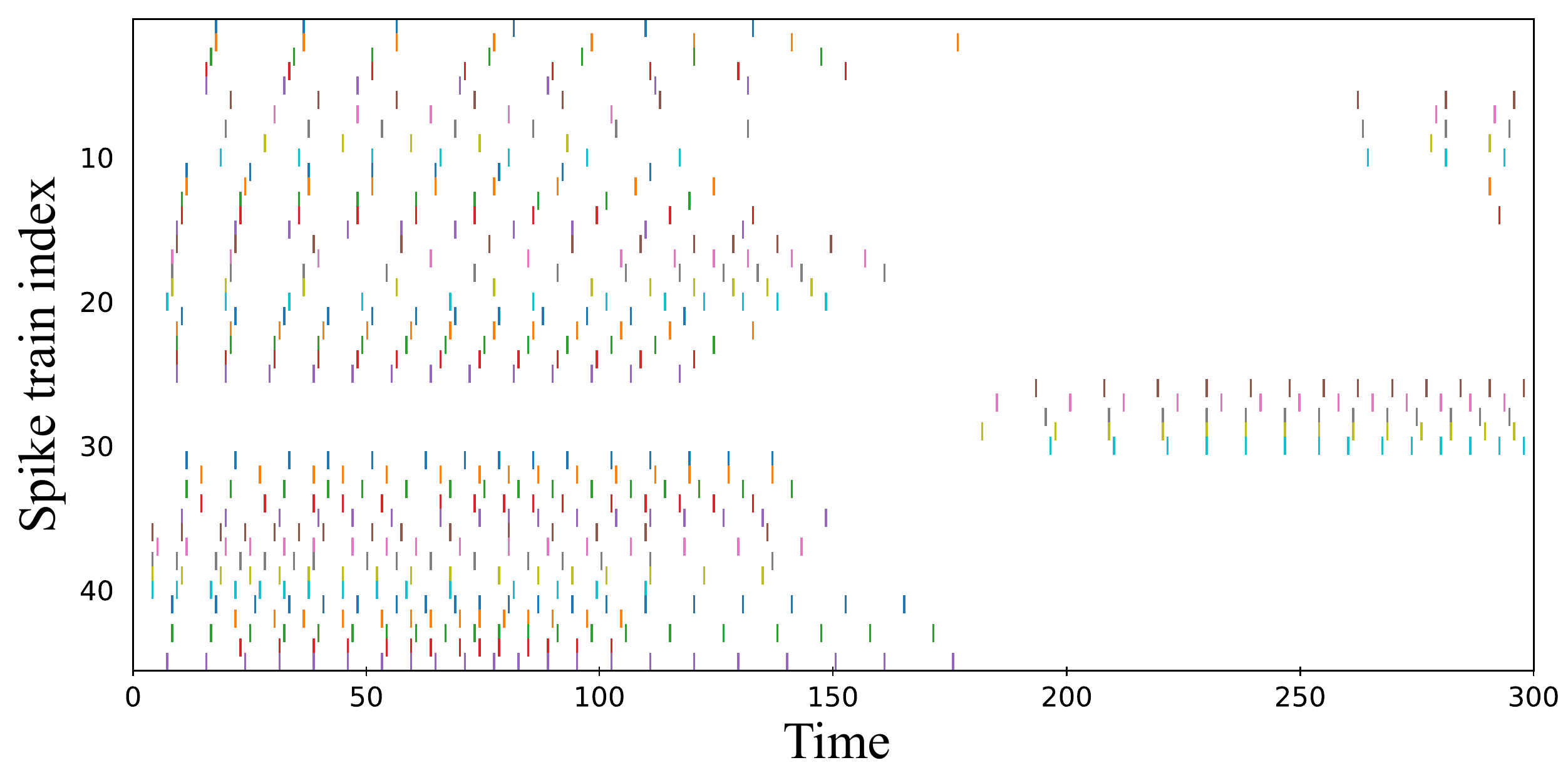}
    \caption{Temporal population coding}\label{fig:temporal}
  \end{subfigure}
  
  \caption{Coding comparison}
  \label{fig:coding_compare}
\end{figure}

First, we study the efficiency of the proposed coding method by utilizing the Articulary Word Recognition dataset collected by UEA \& UCR Time Series repository \cite{bagnall2018uea} as an example. This dataset consists of multivariate sensor data recorded by Electromagnetic Articulograph (EMA), which is a device to track the motion of speakers' tongue and lips. Each sample contains 9 variates of length 144. An example of this dataset is shown in figure \ref{fig:sample}, each line represents a time varying input. We use both rate coding and temporal population coding to convert the time series to spikes. The spike patterns are shown in figure \ref{fig:coding_compare}. Each dot in figure \ref{fig:rate} represents a spike, and in figure \ref{fig:temporal} a spike is represented by a vertical line. We use an input window of 300. Due to the incapability of rate coding to represent temporal information, the time series have to be flattened, resulting in 1296 spike trains. For clarity, only the first 100 spike trains are shown in figure. In temporal coding, for each variate, we use 5 neurons to encode. It is clearly seen that the temporal population coding is sparser. In addition, it is encoding the input with 45 spike trains, which is significantly smaller than the number of spike trains obtained by rate coding. This is beneficial to reduce the SNN model size.

We tested our coding method with rate coding on four multivariate datasets, details including average spike count, spike rate, input size are shown in table \ref{Tab:coding}. Temporal in table \ref{Tab:coding} refers to temporal population coding. The spike rates are significantly lower than rate-based coding. As can be seen in last column, the input size of our coding method is also significant less. Particularly for long time series, such as Atrial Fibrillation dataset. It consists of two variates, and the length is 640, therefore flattening operation resulting a large number of inputs. Buffering such long time series also causes significant latency in real time applications. While temporal population coding can convert input on the fly, not only input size is reduced, buffering is also no longer required.

\begin{table}[h]
\centering
\captionof{table}{Coding Efficiency \label{Tab:coding}}
\begin{tabular}{|c|c|c|c|c|}
\hline
Dataset &
  \begin{tabular}[c]{@{}c@{}} \textbf{\thead{Coding \\ Method}} \end{tabular} &
  \begin{tabular}[c]{@{}c@{}} \textbf{\thead{Spike \\ Count}}\end{tabular} &
  \begin{tabular}[c]{@{}c@{}} \textbf{\thead{Spike \\ Rate}}\end{tabular} &
  \begin{tabular}[c]{@{}c@{}} \textbf{\thead{Input \\ Size}} \end{tabular} \\ \hline
\multirow{2}{*}{\begin{tabular}[c]{@{}c@{}}Articulary Word\\  Recognition\end{tabular}} & Rate     & 65653.8 & 16.90 & 1296 \\ \cline{2-5} 
                                                                                        & Temporal & 518.7   & 3.84  & 45   \\ \hline
\multirow{2}{*}{\begin{tabular}[c]{@{}c@{}}Basic\\ Motions\end{tabular}}                & Rate     & 15061.7 & 8.36  & 600  \\ \cline{2-5} 
                                                                                        & Temporal & 122.4   & 1.13  & 30   \\ \hline
% \multirow{2}{*}{\begin{tabular}[c]{@{}c@{}}UWave\\ Gesture Library\end{tabular}}        & Rate     & 46918.6 & 16.54 & 945  \\ \cline{2-5} 
%                                                                                         & Temporal & 382.5   & 8.49  & 15   \\ \hline
\multirow{2}{*}{\begin{tabular}[c]{@{}c@{}}Finger\\ Movements\end{tabular}}             & Rate     & 1069.0  & 25.45 & 1400 \\ \cline{2-5} 
                                                                                        & Temporal & 471.3  & 1.12  & 140  \\ \hline
\multirow{2}{*}{\begin{tabular}[c]{@{}c@{}}Atrial\\ Fibrillation\end{tabular}}          & Rate     & 23041.4 & 10.77 & 1280 \\ \cline{2-5} 
                                                                                        & Temporal & 164.5   & 4.76  & 10   \\ \hline
\end{tabular}
\end{table}

\subsection{Computation Overhead}

To evaluate the computation overhead of proposed coding method and event driven inference algorithm, we build a SNN to classify Articulary Word Recognition dataset. A vanilla 2 layer stacked LSTM and RNN of unit size 300 are also built as reference. The network structure, number of network parameters, and accuracy are shown in table \ref{Tab:comp_overhead}. Our network achieved comparable accuracy with 11 \% number of parameters of LSTM. In addition, the length of this time series is 144, LSTM and RNN have to perform computation step by step, this introduces significant amount of operations. Our model can benefit from the event driven nature, computations are only necessary when there are spike events. Given the average number of input spike is 518.7, the computation overhead is minimal compared with LSTM/RNN.

\begin{table}[h]
\centering
\captionof{table}{Model comparison\label{Tab:comp_overhead}}
\begin{tabular}{|c|c|c|c|}
\hline
\textbf{Model}  &  \textbf{Network structure} &  \textbf{Parameter number} & \textbf{Accuracy}   \\ \hline
LSTM    & 9-300-300-25  &   1103125   &     98.31          \\ \hline
RNN     & 9-300-300-25  &   281425 &        98.20        \\ \hline
SNN     & 45-300-300-25 &   125880  &        98.27      \\ \hline
\end{tabular}
\end{table}

\subsection{Time Series Classification}

\begin{table}[h]
\centering
\captionof{table}{Accuracy \label{Tab:acc}}
\begin{tabular}{|c|c|c|c|c|c|}
\hline
Dataset             & \thead{ED \\ \cite{bagnall2018uea}}    &  \thead{DTW \\ \cite{bagnall2018uea}}   & \thead{TapNet \\ \cite{zhangtapnet} }   &\thead{WEASEL \\ \cite{schafer2017multivariate} }   & \thead{This \\ work }\\ \hline
\begin{tabular}[c]{@{}c@{}}Articulary Word \\ Recognition\end{tabular}      & 0.97  & 0.98  & 0.987  & -     &    0.98   \\ \hline
% \begin{tabular}[c]{@{}c@{}}UWave Gesture\\ Library\end{tabular}  & 0.881 & 0.869 & 0.894  & -     &   0.89    \\ \hline
\begin{tabular}[c]{@{}c@{}}FaceDetection\end{tabular}  & 0.519 & 0.513 & 0.556  & -     &   0.57    \\ \hline
BasicMotions        & 0.675 & 1     & 1      & -     &   1       \\ \hline
% Libras              & 0.833 & 0.894 & 0.85   & 0.894 &   0.845     \\ \
Heartbeat              & 0.62 & 0.659 & 0.751   & - &   0.72     \\ \hline
\begin{tabular}[c]{@{}c@{}}Spoken Arabic\\ Digits\end{tabular}  & 0.967 & 0.96  & 0.983  & 0.992 &   0.98     \\ \hline
% PenDigits           & 0.973 & 0.939 & 0.98   & 0.912 &    0.974    \\ \hline
JapaneseVowels           & 0.924 & 0.959 & 0.965   & 0.976 &    0.97    \\ \hline
RacketSports        & 0.868 & 0.842 & 0.868  & 0.934 & 0.87  \\ \hline
\end{tabular}
\end{table}

Our algorithm is evaluated on 7 multivariate time series datasets. We build a network of size 500-500-500-X, X indicates the size of last layer, which varies according to different dataset class numbers. Adam optimizer is used and the learning rate is 0.0001. The result is shown in table \ref{Tab:acc}. ED and DTW refer to 1-Nearest Neighbor with Euclidean Distance and Dynamic Time Warping respectively \cite{bagnall2018uea}. TapNet is a DNN-based approach for time series classification \cite{zhangtapnet}. No accuracy in SNN domain is listed, to our best knowledge, there is no previous work that comprehensively focusing on time series classification with SNN.  

In all the 7 datasets, our method outperforms the 1-Nearest Neighbor classifier, which is a standard classifier for time series classification. In Spoken Arabic Digits dataset and Racket Sport dataset, our method achieves higher accuracy than DNN based approach. In Articulary Word Recognition dataset, Heart Beat dataset, thought TapNet achieves better accuracy, however the advantage is insignificant: 0.987 v.s. 0.98, 0.751 v.s. 0.72.

\section{Conclusion}

In this work, we presented an iterative SNN model and training algorithm for spatial temporal spike pattern classification. A coding method to convert continuous time series to discrete spikes is also proposed. Our coding method is able to represent information by sparse spike patterns, such that the computation overhead can be significantly reduced. We evaluate our algorithm and coding method on various multivariate time series dataset, and outperform the standard 1-Nearest Neighbor classifiers and also show competitive performance with DNN based approaches.

\bibliographystyle{abbrv}
\bibliography{bib}

\end{document}